\documentclass[lettersize,journal]{IEEEtran}
\usepackage{amsmath,amsfonts}
\usepackage{algorithmic}
\usepackage{algorithm}
\usepackage{array}
\usepackage[caption=false,font=normalsize,labelfont=sf,textfont=sf]{subfig}
\usepackage{textcomp}
\usepackage{stfloats}
\usepackage{url}
\usepackage{verbatim}
\usepackage{multirow}
\usepackage{graphicx}
\usepackage{cite}
\begin{document}

\title{Weakly-Supervised Cross-Domain  Segmentation of Electron Microscopy with Sparse Point Annotation}

\author{Dafei Qiu, Shan Xiong, Jiajin Yi, Jialin Peng*
\thanks{D. Qiu, S. Xiong, and J. Yi contributed equally. This work was supported in part by  National Natural Science Foundation of China (No. 11771160 and No. 62276105),  Natural Science Foundation of Xiamen (No. 3502Z202373042).}
\thanks{The authors are with the College of Computer Science and Technology, Huaqiao University, Xiamen 361021, China.}
\thanks{* Corresponding author: J. Peng (E-mail: 2004pjl@163.com).}}

\markboth{IEEE Transactions on Big Data}%
{Shell \MakeLowercase{\textit{et al.}}: A Sample Article Using IEEEtran.cls for IEEE Journals}


\maketitle

\begin{abstract}
Accurate segmentation of organelle instances from electron microscopy (EM) images plays an essential role in many neuroscience researches. However, practical scenarios usually suffer from high annotation costs, label scarcity, and large domain diversity. While unsupervised domain adaptation (UDA) that assumes no annotation effort on the target data is promising to alleviate these challenges, its performance on complicated segmentation tasks is still far from practical usage. To address these issues, we investigate a highly annotation-efficient weak supervision, which assumes only sparse center-points on a small subset of object instances in the target training images. To achieve accurate segmentation with partial point annotations, we introduce instance counting and center detection as auxiliary tasks and design a multitask learning framework to leverage correlations among the counting, detection, and segmentation, which are all tasks with partial or no supervision. Building upon the different domain-invariances of the three tasks, we enforce counting estimation with a novel soft consistency loss as a global prior for  center detection, which further guides the per-pixel segmentation.  To further compensate for annotation sparsity, we  develop a cross-position cut-and-paste for label augmentation and an entropy-based  pseudo-label selection. The experimental results highlight that, by simply using extremely weak annotation, e.g., 15\% sparse points, for model training, the proposed model is capable of significantly outperforming  UDA methods and produces comparable performance as the supervised counterpart. The high robustness of our model shown in the validations and the low requirement of expert knowledge for sparse point annotation further improve the potential application value of our model. Code is available at https://github.com/x-coral/WDA.
\end{abstract}


\section{Introduction}
\begin{figure}[t]
\centering
\includegraphics[width=0.49\textwidth]{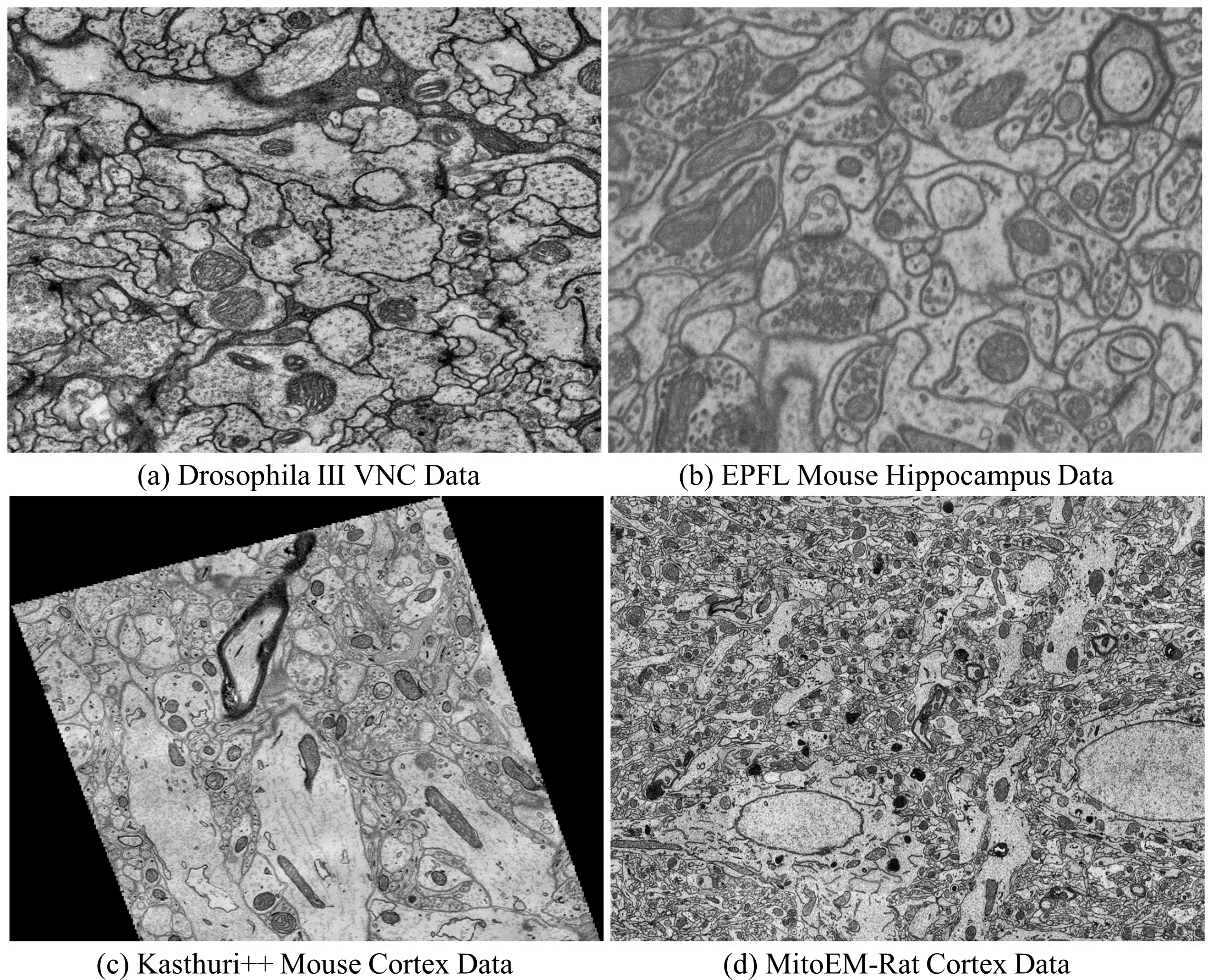}
\caption{Typical EM images scanned from different tissues/species, which result in large domain gaps.   } \label{fig:1}
\end{figure}

\begin{figure*}[tb]
\centering
\includegraphics[width=0.95\textwidth]{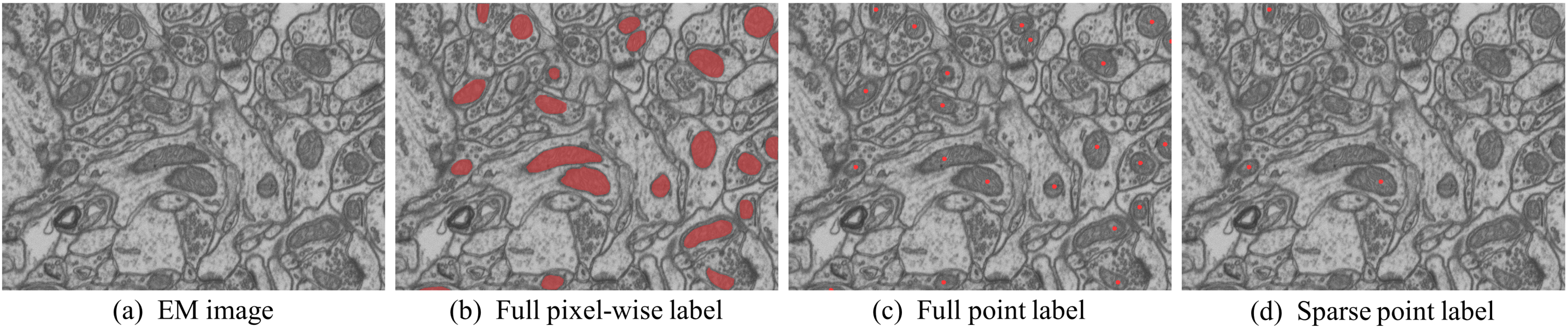}
\caption{Illustration of the sparse point annotation and other types of annotations for mitochondria in EM images. Sparse point annotation not only leads to a significantly reduced annotation workload but also requires much less expert knowledge.
} \label{fig:2}
\end{figure*}

\IEEEPARstart{T}{he}
 increasing  acquisition ability of electron microscopy (EM), such as  serial section electron microscopy (SEM) and scanning transmission electron microscopy (sTEM),  enables 3D quantifying ultrastructure and morphological complexities of cell organelles, e.g., mitochondria,  in nanometre-level, which is essential for advancing our understanding of cell biology and various diseases \cite{nunnari2012mitochondria,neikirk2023call,pekkurnaz2022mitochondrial}.
For example, Liu et al. \cite{liu2022fear} found that fear conditioning significantly increases the number of mitochondria but decreases their size, yielding insight into cell plasticity  associated with fear learning.
However, mitochondrial quantification through manually
annotating a large number of mitochondria instances in large-scale volume electron microscopy (EM) is extremely time-consuming and usually prohibitive. Thus, there is an increasing demand for automatic  segmentation of  subcellular organelles, e.g., mitochondria as shown in Fig. \ref{fig:1} and Fig. \ref{fig:2}, which have intrinsic linkage with cellular homeostasis, function, and numerous diseases \cite{nunnari2012mitochondria}.

Recently, supervised  models \cite{ronneberger2015u,shelhamer2017fully,peng2021medical} have had remarkable success on various  segmentation tasks including  mitochondria segmentation.  However, the high performance significantly degrades when the labeled training data are insufficient  or the testing data are drawn from different distributions.
The high annotation cost for biomedical images and  widespread data distribution shifts usually  preclude their applications in practical scenarios, because fully supervised models not only  highly rely on a large amount of per-pixel labeled data for model training  but also are very sensitive to data distribution shift (also known as \textit{domain shift}). For large-scale EM volumes, per-pixel  labeling of tiny sub-cellular organelles by experts is  time-consuming, labor-intensive, and also expensive since there are  plenty of mitochondria instances with varied shapes, sizes, and appearances as well as ambiguous boundaries on each slice of EM volumetric image (Fig. \ref{fig:2}). Moreover, due to the heterogeneity of the cells and the varied imaging methods, EM images obtained from different tissues/species with different types of electron microscopes show significant content and appearance differences, which present as significant domain shifts (Fig. \ref{fig:1}). While fully supervised training on each domain will produce models with leading performance, collecting sufficient high-quality labeling by manual annotation is usually prohibitive and especially daunting for various EM images.

 A promising solution that can alleviate the domain shift and lighten  the heavy annotation burden  on a new target domain is to conduct domain adaptation (DA), which explores  a related but well-annotated  source domain and  bridges the domain gap by learning domain-transferable representations \cite{DAN}. In recent years,  unsupervised domain adaptation (UDA), which uses no labels on  target data, has been extensively studied and  has also achieved impressive progress \cite{DANN,Outputspace,peng2020unsupervised} on various tasks. However, due to the completely missing supervision on the target domain, there are still  huge performance gaps between  UDA methods and  fully supervised counterparts, especially for  high-dimensional prediction tasks such as image segmentation. Especially, UDA approaches are  far from practical usage for challenging biomedical segmentation tasks due to their relatively low performance.  Although a simple and natural idea to boost  model generalization ability is to densely label a subset of target images\cite{chen2021semi}, this semi-supervised domain adaptation (SDA)  requires significantly increased annotation cost, time effort, and domain knowledge by experts. Thus, we propose to use a type of extremely weak and incomplete annotation, which can significantly reduce the annotation burden and also require a much lower level of expert knowledge. Moreover, we have observed that pixel-wise annotations usually contain redundancy, especially  in EM images that contain many instances of various subcellular organelles.  For practical usage, it is crucial  to address the annotation bottleneck and performance limitation in cross-domain model adaptation.

In this work,  we introduce a novel class of \textit{weakly-supervised domain adaptation (WDA)} setup to achieve a high-performing model  with minimal annotation cost at the training stage. Instead of using fully pixel-wise  annotation of the EM images as in Fig. \ref{fig:2} b,  we assume that the  training data of the target domain have sparse center-point annotations on a randomly-sampled  small subset (e.g., 15\%) of mitochondria instances (Fig. \ref{fig:2} d), which can easily be accomplished by non-experts in a very short time. In particular, labeling ambiguous boundaries and  distinguishing hard instances can be avoided in our sparse point annotation.  While  scribbles \cite{gao2022segmentation,liu2022weakly} have been widely used as sparse annotations for medical image segmentation, this annotation type still requires great annotation effort and is less suitable for segmenting plenty of small objects. Our sparse point annotation is essentially an extreme case of scribbles with minimal annotation area and is a more determined process than the scribbles. Even compared to full center-point annotation (Fig. \ref{fig:2} c),  the proposed \textit{sparse point annotation} requires significantly reduced annotation time and domain expertise (see Sec. \ref{sec:result} for quantitative comparison).  However, giving center locations of a small proportion of instances and leaving most pixels unlabeled results in a very  challenging setting.
Compared to other weak labels \cite{peng2021medical}, e.g., bounding boxes and full center points, the proposed sparse point annotation just provides location information of a small proportion of foreground object instances in the target domain but without any boundary and appearance information about these instances. Thus, another obstacle to training the segmentation model with sparse point labels  lies in that  no background label is available and the unlabeled pixels contain both foreground objects and true background.

Thus, we propose a novel multi-task pyramid learning framework, namely WDA-Net to  perform cross-domain segmentation of organelle instances from electron microscopy  with extremely sparse point annotations on the target training data.  Given the weak and incomplete supervision, we introduce center detection and counting as two auxiliary  tasks for the segmentation and jointly learn the three correlated tasks for multi-level domain alignment.  Intuitively, the segmentation of plenty of object instances will be simplified when we have center-point locations;  the center-point detection task will be simplified when we know the total number of object instances. Since  these three tasks have only  partial or no supervision, the proposed WDA-Net further takes advantage of the different levels of domain invariance of them. One observation is that, given the  label information on a related source domain, roughly counting and locating  similar objects in an unlabeled target domain is much easier than accurately segmenting them. In fact, the counting task is usually easier than center detection as observed by many studies\cite{zhang2018crowd}. Thus, at the training stage, the WDA-Net leverages the predictions of a counting network trained on the source domain as a soft  global prior for the  weakly-supervised center-detection  task, which further provides location information to the segmentation task. Moreover, we bridge the detection  and segmentation tasks  by using shared semantic features. To further alleviate the annotation sparsity, we introduce a cross-location cut-and-paste  augmentation to increase the density of annotated points.  Three challenging datasets are used for model validation and comparison.  In summary, our main contributions are as follows,
\begin{itemize}
    \item  We introduce a highly annotation-efficient  type of weak supervision, sparse center-points on partial object instances of target training images, for the accurate cross-domain segmentation of cellular images.
    \item Given the extremely weak and incomplete supervision, we  introduce counting and center detection as auxiliary tasks and  design a multi-task learning framework, namely WDA-Net, to take advantage of the correlations among different tasks.
    \item A novel soft consistency loss is introduced to constrain the detection with estimated global counting prior.
     \item Cross-location cut-and-paste label augmentation and entropy-based pseudo-label selection are developed to alleviate the annotation sparsity.
      \item Experiments on challenging tasks show that the proposed WDA-Net significantly outperforms SOTA UDA counterparts and shows performance close to the fully supervised model.
\end{itemize}

This study is a substantial extension of our preliminary conference version \cite{qiu2022wda} with  more comprehensive validations on more datasets, more detailed model analysis and comparisons,  as well as a  more comprehensive literature review and more detailed concept illustrations. Moreover, a label augmentation named source annotation refinement (SAR) discussed in Sec. \ref{sec:result}  is introduced to address the domain gap resulting from  imperfect boundary annotation of the source data.

The remainder of this paper is organized as follows. Sec. \ref{sec:relatedwork} gives a brief review of related work, and Sec. \ref{sec:method}  provides the details of our method. Experimental results are presented in Sec. \ref{sec:result}.  Finally, the study is concluded  in Sec. \ref{sec:conclusion}.

\begin{figure*}[t]
\centering
\includegraphics[width=0.99\textwidth]{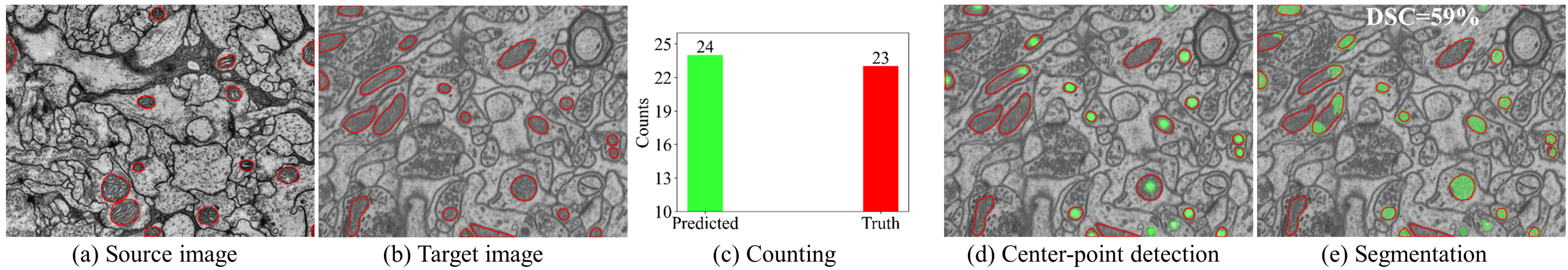}
\caption{Performance of source models without domain adaptation on the target domain. Red: ground truth annotations; Green: predicted center points/segmentation. a) A source image from Drosophila III VNC data \cite{DrosophilaIIIVNC}; b) a target image from EPFL data \cite{lucchi2013learning}; c) predicted counts by the source counting model; d) predicted center-points by the source detection model; e) predicted segmentation by the source segmentation model.   } \label{fig:3}
\end{figure*}

\section{Related Work} \label{sec:relatedwork}
\textbf{Biomedical image segmentation.}  Recently, microscopy image segmentation  has attracted much research attention  \cite{xing2016robust,raza2019micro,lucchi2013learning,peng2021csnet,peng2019mitochondria}. Compared to classical machine-learning based models \cite{lucchi2013learning,yuan2021hive,peng2019mitochondria},   deep fully convolutional networks (FCNs), especially U-Net and its variants, have shown state-of-the-art (SOTA) performance\cite{peng2021csnet,casser2020fast} on many tasks, including EM image segmentation.  For instance, an improved U-Net  is developed by \cite{casser2020fast} for  mitochondria recognition and segmentation  in EM images, while  a deeply supervised 3D  U-Net with residual connections is utilized by \cite{xiao2018automatic} for  mitochondria segmentation. To adapt to the limited computation resources in practical scenarios, a lightweight 2D encoder-decoder named CS-Net is proposed   in our previous study \cite{peng2021csnet}  and showed SOTA  performance for segmenting mitochondria from EM images and segmenting nuclei  from histology images.  To learn diverse contexts and lighten the deep networks,  the CS-Net replaces the expensive standard convolutions with lightweight hierarchical dimension-decomposed convolutions (HDD).  Given its sound performance,  we use the HDD module as the basic building block of our WDA-Net.

\textbf{Weakly-supervised segmentation.}
Recently, there has been  increasing attention to the exploration of various weak labels \cite{peng2021medical}, e.g., image-level label \cite{2018Learning}, bounding-box \cite{2021Box,xie2022wits},  scribbles \cite{dorent2020scribble,liu2022weakly},  and points \cite{Nishimura2021,2020Weakly,obikane2019weakly,bearman2016s},  to alleviate the heavy annotation burden. In the field of  weakly-supervised learning,  full point annotation \cite{Nishimura2021,2020Weakly}  has been used  in cell image segmentation to reduce the annotation burden.  For nuclei segmentation in digital pathology, \cite{tian2020weakly} utilized point annotation and  used  a self-supervision strategy with clustering to obtain coarse segmentation.  \cite{hui2020weakly} also exploited  partial point annotation to segment  nuclei from  histopathology images, and extracted pseudo labels from Voronoi partition and k-means clustering. A densely-connected conditional random field (CRF) is employed to refine the coarse segmentation. In contrast to the nuclei that are densely distributed
 in histopathology images, mitochondria are usually  sparsely distributed in nano-scale EM images, which makes the Voronoi partition and clustering less efficient. \cite{chen2021weakly} assumed points on both the foreground  and  background and  formulated  the task as a partially-supervised super-pixel classification  problem.  In contrast to the  nuclei in histopathology images, mitochondria  in EM images are usually  sparsely distributed with largely varied  shapes, making Voronoi partition, clustering, and super-pixel segmentation  less effective. In contrast to these studies, we investigate domain-adaptive segmentation of EM images  with few center points as supervision on target training images.

\textbf{Cross-domain segmentation.}
As a typical type of DA, UDA assumes no annotation on the target domain and thus has been a prominent  problem setting in many tasks \cite{dong2022unsupervised,do2022exploiting,gao2022cross}.  Given the great success of  UDA methods in many classification tasks \cite{DANN}, they have also been applied in segmentation tasks, which involve more challenging  structured prediction. Popular methods usually seek to reduce the domain discrepancy by minimizing distribution discrepancy loss \cite{DAN}, or conducting  reconstruction learning\cite{DeepG},  style transfer \cite{hoffman2018cycada}, and adversarial alignment \cite{DANN,Outputspace}. Although style transfer \cite{hoffman2018cycada} with adversarial image-to-image translation can explicitly reduce distribution differences of source and target images in pixel and other low-level space, they usually fail to incorporate high-level semantic knowledge. Thus, a cascade of style transfer and other domain alignment is usually used \cite{hoffman2018cycada}.
 For cross-domain EM image segmentation,  \cite{Y-Net} introduced the Y-Net, which learns shared features that can reconstruct  images from both the source  and target domains.
As a typical type of adversarial-alignment-based method, the  AdaptSegNet introduced by  \cite{Outputspace} aligns the distributions of the source and target domains in both output space and low-level feature space with multi-level adversarial learning and has achieved SOTA  performance in many tasks.
The DAMT-Net in \cite{peng2020unsupervised}, which has achieved top performance for the cross-domain mitochondria segmentation, explores transferable geometrical cues and visual cues and  conducts  domain alignment on multiple feature spaces with both reconstruction learning and adversarial alignment. Rather than only conducting slice-by-slice 2D segmentation, the top-performing DA-ISC \cite{huang2022domain} utilized 2.5D input and enforced inter-section consistency. The predicted inter-section residuals and segmentation of source and target volumes are aligned via adversarial training.

Another category  of methods that do not directly minimize domain discrepancy is to conduct  pseudo-label learning on the domain data \cite{2018Domain} with the source model. Despite  the promising performance, methods based on self-training  rely heavily on the performance of the source model and the strategy of confident pseudo-label selection.  A SOTA method of this class is the SAC \cite{araslanov2021self}  that uses  self-supervised augmentation consistency and co-evolving pseudo labeling.

Despite the impressive progress of UDA methods, their performance is
still much lower than supervised methods, especially on complicated segmentation tasks. Recently, researchers also have considered various forms of weak annotations. For domain-adaptive multi-class segmentation tasks,  \cite{Paul2020ECCV} considered image-level labels and category-wise point labels (i.e., one point for each class on each image), which specify the categories that occur in each target image. However, category information provided by image-level annotation and category-wise point labels  is less informative for our binary mitochondria segmentation task, which involves  delineating a large number of object instances.
\cite{2021Box} investigated bounding-box annotations as weak supervisions on the target domain and has obtained  impressive results for domain-adaptive  liver segmentation. However, it is also  time-consuming and laborious to manually label  bounding boxes for a large number of  small organelles in large-scale EM images.  To achieve an annotation-efficient method, we utilize  center-point labels on partial object instances, which are much more efficient to label, even by annotators with little domain expertise. To the best of my knowledge, this study is the first  study that exploits sparse center-point labels for domain-adaptive segmentation.

\begin{figure*}[t]
\centering
\includegraphics[width=0.85\textwidth]{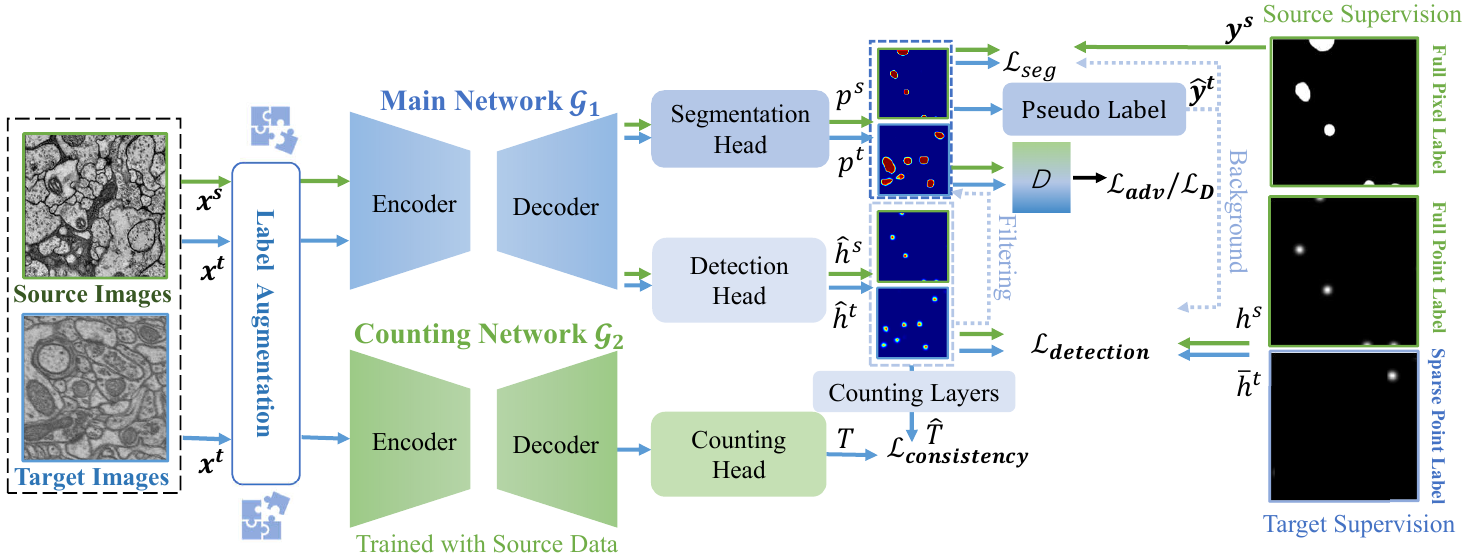}
\caption{Illustration of our proposed WDA-Net, which jointly learns
 three correlated tasks that have different levels of domain invariance. An auxiliary counting task is employed to guide  the detection task, which helps  recognize mitochondria instances and reduce false positives in both implicit and explicit ways.
  } \label{fig:4}
\end{figure*}
\section{Methodology} \label{sec:method}
To achieve a high-performing but annotation-efficient method for domain-adaptive segmentation of images having plenty of object instances, we consider weakly-supervised domain adaptation with sparse center-point annotations on target training data. Let $\mathcal{D}_s$ be the set of source images with full pixel-wise annotation $y^s\in \{0,1\}^{H,W}$ for each source image $x^{s}$ of size $H\times W$, as illustrated in Fig. \ref{fig:2} (b). Given the full label image $y^s$, we can also obtain an auxiliary full point label $c^s\in \{0,1\}^{H,W}$that only takes 1 at the mass center of each object instance in  $x^{s}$.
Given the target training data $\mathcal {D}_t$, we further assume having access to sparse center-point annotations  $\bar{c}^t \in \{0,1\}^{H,W}$, which takes 1 only on the center of a small subset of foreground object instances as shown in Fig. \ref{fig:2} (d).  Typically, a well-trained model on the source data $\mathcal{D}_s$ severely degrades on the target data due to the domain gap. Our goal is  to learn a high-performing model for target data with extremely weak annotation only at the meta-training stage.

As demonstrated in Fig. \ref{fig:4}, our WDA-Net is formulated as a multi-task learning framework that jointly learns  three highly correlated tasks, i.e., counting, center detection, and segmentation, for multi-level domain alignment. The proposed WDA-Net is comprised of two sub-networks:  the $\mathcal{G}_1$  network used for center detection and segmentation,  and an auxiliary network $\mathcal{G}_2$ for the counting task. While the $\mathcal{G}_1$  is trained using both the source images and target images, the auxiliary counting network $\mathcal{G}_2$  is  trained only using the source data.

The $\mathcal{G}_1$  has two prediction heads for segmentation and detection, respectively,  but with a shared encoder-decoder for joint learning. For each source/target input image $x$, the segmentation head predicts a segmentation probability map $p$, and the detection head outputs a heatmap $\hat{h}$ that  takes peaks at centers of object instances. The main challenges lie in that both the segmentation and detection tasks have incomplete and insufficient supervision.  Given the large domain gap for cross-domain dense segmentation,  we guide the segmentation process by pseudo-label learning and adversarial learning, which is presented in Sec. \ref{subsec:segmentaton}. The predicted centers by the detection head are also used to remove false positive segmentation. The center detection presented as a heatmap regression problem is guided by the partial supervision  and global prior provided by the counting network, which are presented in Sec. \ref{subsec:detection}. Note that the sparse point annotations provide much more effective supervision to the center detection than that to the dense segmentation.

In contrast to the detection network, the counting network $\mathcal{G}_2$ outputs instance counts, which is used to constrain the detection, especially at the early training stage. Despite the domain gap, we observed that the cross-domain counting task is much easier than the dense segmentation and center  detection. A simple empirical investigation   is shown in Fig. \ref{fig:3}.  With proper data augmentation, the counting model shows higher domain invariance. Given the inaccuracy of the counting prediction, we propose a novel consistency loss with a soft margin to constrain  the detection with instance counts.

 \subsection{Segmentation Head}\label{subsec:segmentaton}
Given the domain gap, we utilize adversarial learning \cite{2018Domain,Outputspace} to learn domain-invariant semantic features shared by the segmentation and center detection. Moreover, we explore both the ground truth annotations  of source images and  pseudo-labels generated for the target training images to learn the segmentation head. Let  $L_{ce}(\cdot,\cdot)$ refer to the cross-entropy cost function,  the segmentation loss is defined  as follows,
\begin{equation}\label{eq:1}
\begin{aligned}
\mathcal{L}_{seg}=\frac{1}{|\mathcal{D}_s|}\sum_{x^s} L_{ce}(p^s,\mathbf{y}^s)+\frac{1}{|\mathcal{D}_t|}\sum_{x^t} L_{ce}(p^t,\hat{\mathbf{y}}^t)
\end{aligned}
\end{equation}
in which $p^s$ and $p^t$ refer to  the predicted probability maps of the source and target segmentation $p(x^s)$ and $p(x^t)$, respectively; $\mathbf{y}^s\in \{0,1\}^{H,W,2}$denotes the one-hot encoding mask of the binary label image $y^s$; $\hat{\mathbf{y}}^t$ denotes the partial one-hot encoding mask  of the pseudo-label image  for $x^t$ and takes $\textbf{0}$ on unlabeled pixels, which will be ignored in gradient back-propagation of network optimization.

\textbf{Pseudo-label generation.}
To improve the cross-domain segmentation, we conduct self-training (as described in Eq. \ref{eq:1}) by exploring pseudo-labels on unlabeled pixels of the target training images. While pseudo-labels generated by selecting the most probable class with predefined threshold \cite{peng2021medical} may be very noisy and overconfident, we take advantage of entropy \cite{saporta2020esl,lee2013pseudo} of the softmax predictions and devise an entropy-based pseudo-label selection method. Let $p^{t}_{i,l}$ be the abbreviation of $p(x^t)_{i,l}$, then we can compute the  pseudo-label $\hat{\mathbf{y}}^{t}$ at the $i$th pixel for the class $l$  as follows,
\begin{equation}
\hat{\mathbf{y}}_{i,l}^{t}=\left\{\begin{array}{lc}
1, &\mathrm{if} ~l=\arg\max_{\tilde{l}}~ p^{t}_{i,\tilde{l}}\ \mathrm{and}\ E(p^{t}_{i})<v_{l} \\
0, &\rm{otherwise}
\end{array}\right.
\end{equation}
in which   $p^{t}_{i}=[p^{t}_{i,1}, p^{t}_{i,2}, \cdots, p^{t}_{i,L}]^T$, $E(\cdot)$ refers to the normalized entropy function.
\begin{equation}
 E(p^{t}_{i})= -\frac{1}{\log(L)}\sum_{l=1}^Lp^{t}_{i,l}\log p^{t}_{i,l}
\end{equation}
To obtain reliable pseudo-labels, we use $v_l$ as a threshold over the entropy score for the $l$ th class:
\begin{equation}
v_l= D_K \{E(p^{t}_{i}) \mid x^{t}_i \in \mathcal {D}_t, l=\mathrm{\arg \max}_{\tilde{l}} ~p^{t}_{i,\tilde{l}}\}
\end{equation}
in which $D_K$ refers to the $K$th decile. By setting  $K$=8, we select  the top 80\% most confident predictions as pseudo labels.  The self-training process dynamically updates the label status of unlabeled pixels in the next training stages.

\textbf{Adversarial learning.}
To tackle the adaptation of the source segmentation model,  we use adversarial learning as many previous studies \cite{DANN,Outputspace} and explore the similarity between  the segmentation spaces of the source domain and target domain. Thus, following the idea in \cite{Outputspace}, we enforce  a fully convolutional discriminator network $D$ on the predictions of the segmentation head to  discriminate whether the  inputted segmentation  is  from the source prediction or the target prediction. Meanwhile, the segmentation generator, i.e., the segmentation network competes against  the discriminator. Thus,  an  adversarial loss $\mathcal{L}_{adv}$ is enforced to train the segmentation network and fool the discriminator.     While we can conduct the adversarial learning in other feature spaces \cite{DANN,peng2020unsupervised}, we choose to align the label distribution by taking advantage of the spatial layout and instance-level shape similarity between the source and target domains.
 For model training, we  train the domain discriminator ($D$) with the following objective function, $\mathcal{L}_{D}$  \cite{Outputspace},
  \begin{equation}
\mathcal{L}_{D}=-\frac{1}{|\mathcal{D}_s|}\sum_{x^s}\log D(p^{s})- \frac{1}{|\mathcal{D}_t|}\sum_{x^t}\log D(1-p^{t})
\end{equation}
  The generator  $\mathcal{G}_1$ is trained by jointly optimizing  the segmentation loss $\mathcal{L}_{seg}$ and the  adversarial loss $\mathcal{L}_{adv}$  \cite{Outputspace},
  \begin{equation}
\mathcal{L}_{adv}=- \frac{1}{|\mathcal{D}_t|}\sum_{x^t}\log D(p^{t})
\end{equation}
 which is tasked to confuse the discriminator.
With adversarial learning, we achieve model adaptation through matching the prediction distributions. Thus, we can also obtain an adapted  model for estimating pseudo-labels.
\subsection{Detection Head with Partial Point Supervision} \label{subsec:detection}
To boost the learning of the segmentation network,
we utilize multitask learning and introduce an auxiliary center-detection head (as shown in Fig. \ref{fig:4}), which directly predicts  heatmaps for  mitochondria locations. While the partial center points on the target images provide almost negligible information for the segmentation, this point annotation is much more useful for  center detection. Given  sparse center-point labels $\left\{\bar{c}^t\right\}$ for   target images, the center-detection is a standard regression problem with incomplete annotation. Let $h^{s}$ and $\bar{h}^{t}$  be the ground truth heatmaps ($h^{s}=G_{\sigma_1}*c^s$, and $\bar{h}^{t}=G_{\sigma_1}*\bar{c}^t$) for the source label $c^s$ and target weak label $\bar{c}^t$, respectively,
where $G_{\sigma_1}$ denotes a normalized Gaussian kernel with  bandwidth $\sigma_1$. Thus, summing the density map can produce the point counts in the label image $c^s$/$\bar{c}^t$.  While taking zero in the $h^{s}$ means the background, regions taking zeros in $\bar{h}^{t}$ contain both background and unlabeled foreground instances. Thus,  we introduce the following partial supervision loss that is essentially computed on partial regions in each target training image.
\begin{equation}\label{Eq:6}
\begin{aligned}
\mathcal{L}_{detection}=&\frac{1}{|\mathcal{D}_s|}\sum_{x^s,i}\left(1+\lambda \beta_{i}^{s}\right)\left(\hat{h}_{i}^{s}-h_{i}^{s}\right)^{2}+\\
&\frac{1}{|\mathcal{D}_t|}\sum_{x^t,i} \left(w_{i}+\lambda \beta_{i}^{t}\right)\left(\hat{h}_{i}^{t}-\bar{h}_{i}^{t}\right)^{2}
\end{aligned}
\end{equation}
in which $\hat{h}^{s}$ is the predicted heatmap of  source image $x^s$, and $\hat{h}^{t}$ is the predicted heatmap of target image $x^t$.  For the target training images that are only partially annotated, we introduce a spatial weight map $w$, which takes zeros except for the regions that take  positive values in  $\bar{h}$ and the regions that are definitely predicted as the background under a small threshold $\rho$ (0.1 in the experiments). In other words, the loss $\mathcal{L}_{detection}$ will neglect regions with no knowledge. More specifically, the value of $w$ on pixel $i$ is defined as,
\begin{equation}
w_{i}=\left\{\begin{array}{cc}
1 & ~~p^{x^t}_{i,1}< \rho ~ \mathrm{or}~ \bar{h}_i^{t}> 0 \\
0 & \rm otherwise
\end{array}\right.
\end{equation}
Given the extremely unbalanced foreground and background in $\hat{h}^{s}$ and $\hat{h}^{t}$, we
place a higher weight in a small neighborhood of the labeled center points. To this end, we introduce an additional weight $\lambda \beta$ in Eq. (\ref{Eq:6}) with $\beta^s=G_{\sigma_2}*c^s$ and  $\beta^t=G_{\sigma_2}*\bar{c}^t$. The parameter $\sigma_2$ can control the extent of regions with focus and the positive parameter $\lambda$ controls the magnitude.  Since we only have high confidence in each labeled point and its small neighborhood, we set $\sigma_2$=2  to be smaller than $\sigma_1$=10.

To guide the segmentation with the center detection, we have the segmentation network and center-detection network share most feature layers as shown in Fig. \ref{fig:4}. In this way,  the learning detection task first influences the segmentation  implicitly. Moreover, we use the predicted heatmaps to explicitly help identify confident foreground and background pixels for the segmentation.  Since the peaks of the predicted heatmap indicate the locations of object instances with higher confidence, we utilize the predicted locations and connected component analysis to reduce the false positive segmentation.

\subsection{Counting Head as a Global Prior}
For the considered cross-domain location detection, the used point annotation on partial object instances only can provide  partial information about the foreground objects and leaves the ‘background’ containing both foreground objects and background stuff, which will mislead  the model learning. In other words, the  center-location regression lacks constraints on the target domain.
To this end, we introduce a soft global constraint using an estimated total number of  foreground object instances. While it is unrealistic to learn a counting model on the target domain or accurately estimate the total counts, we utilize a counting model $\mathcal{G}_2$ trained on the source model. Different from the center detection, we directly regress the total number of object instances, which empirically shows higher domain invariance than the location regression and segmentation, as shown in Fig. \ref{fig:3} and Fig. \ref{fig:7}. Thus, the source counting model can provide a rough count of the number of target object instances, which may not be accurate but  useful,  especially  at the early training stage. To this end, we regularize  the center detection  with a novel counting consistency constraint. Let $T(x^t)$ be the estimated  number of object instances for each target training image $x^t$ by the source counting model $\mathcal{G}_2$. Meanwhile, we can also count the number $\hat{T}(x^t)$ of object instances in $x^t$ from the heatmap $\hat{h}^{t}$  predicted by the detection branch of the  network $\mathcal{G}_1$, which can be achieved with  a small counting network or simply an integration layer \cite{zhang2016single}. We enforce consistency between $\hat{T}(x^t)$ and $T(x^t)$ to guide the learning of the center detection model.   Since there is inevitable discrepancy between $T(x^t)$ and  the ground truth count due to inaccurate estimation by $\mathcal{G}_2$, we introduce a soft consistency loss with a small  margin $\varepsilon$ (3 in default in our experiment),
\begin{equation}
\begin{aligned}
&\mathcal{L}_{consistency}=\\&
\frac{1}{|\mathcal{D}_t|}\sum_{x^t}\left\{ \max\{0, \hat{T}-(T+\varepsilon)\}+\max \{0,(T-\varepsilon)-\hat{T}\}\right\}
\end{aligned}
\end{equation}
To improve learning efficiency, the counting model $\mathcal{G}_2$ in encoder-decoder architecture (Fig. \ref{fig:4}) uses the parameters from the source segmentation network as initialization.  Multi-scale input and diverse data augmentation are used to improve cross-domain generalization.

\subsection{Cross-Position Cut-and-Paste Label Augmentation}
To maximally reduce the  cost of manual annotation and the requirement of domain expertise, we propose to use 15\% or less sparse point annotation. However, the extreme sparsity of the partial annotation poses great challenges for both the center detection and segmentation. Moreover, the annotated points are typically unevenly distributed. Inspired by the Cutmix \cite{yun2019cutmix}, we propose to ease the sparsity of the partial point annotation with a \textit{cross-position cut-and-paste augmentation} (CP-Aug) strategy. Let $(x^t_A, \bar{c}^t_A)$ and $(x^t_B, \bar{c}^t_B)$ be  two target training image-label pairs for generating a new image-label pair $(x^t_C, \bar{c}^t_C)$.  From $x^t_A$, we first crop a rectangular region (e.g., 256$\times$256) having the larger number of annotated points  and paste it to a rectangular region  of the same size but with much fewer or no annotation points  in $x^t_B$. Unlike previous methods, we have the cropped  patches not necessarily be pasted  at the  same position.  To further improve the effectiveness of the CP-Aug for sparse-point annotation, we use the estimated pixel-wise labels of the point annotated instances produced by the segmentation head to relabel centers for the cropped piece of the center-labeled object instances near the cropped image boundaries.   In this way, we can avoid misleading the model learning caused by image cropping and have the synthesized images with more annotated points, which can facilitate the model training.

\subsection{Overall Optimization}
  Since we employ adversarial learning \cite{Outputspace} on the segmentation outputs,  we alternatively  minimize a discriminator loss $\mathcal{L}_{D}$  \cite{Outputspace}  and the adversarial loss $\mathcal{L}_{adv}$  \cite{Outputspace} to learn the domain discriminator ($D$) and update $\mathcal{G}_1$, respectively.
More specifically, when the domain discriminator $D$ is fixed, we update the $\mathcal{G}_1$  network through optimize the following objective function,
\begin{equation}
\mathcal{L}_{obj}=\mathcal{L}_{seg}+\lambda_a\mathcal{L}_{adv}+\lambda_d\mathcal{L}_{detection}+\lambda_c\mathcal{L}_{consistency}
\end{equation}
in which $\lambda_a$, $\lambda_d$ are positive weighting coefficients; $\lambda_c$=1-$z$/$z_{max}$ decays as the iteration  $z$, and $z_{max}$  denotes the maximum number of the iteration.

\section{Experiments} \label{sec:result}
\subsection{Dataset and Validation Settings}  Our WDA-Net is evaluated using three challenging datasets, which are produced using different electron microscopes in different resolutions and contain images of various tissues of different species.

\textit{EPFL Mouse Hippocampus Data} \cite{lucchi2013learning} were scanned from the CA1 hippocampus region of a mouse brain using  focused ion beam scanning electron microscope (FIB-SEM) in  an isotropic resolution of 5$\times$5$\times$5 $nm^3$.  This dataset contains two labeled subsets for model training  and testing, respectively. Each image subset has 165 images of size 768$\times$1,024.

\textit{Drosophila III VNC Data} \cite{DrosophilaIIIVNC}  contain 20 images of size 1,024$\times$1,024 taken from Drosophila melanogaster third instar larva VNC using serial section Transmission Electron Microscope (ssTEM). The image stack was scanned at  an in-plane resolution of 4.6$\times$4.6 $nm^2$/pixel and  large slice thicknesses of 45-50 $nm$.

\textit{Kasthuri++ Mouse Cortex Data} \cite{kasthuri2015saturated} were scanned  from  mouse cortex with serial section EM  at a resolution of 3$\times$3$\times$30 $nm^3$. This dataset comprises  a training subset of size 85$\times$1,463$\times$1,613 and a testing subset of size 75$\times$1,334$\times$1,553. We use the  labels provided by \cite{casser2020fast}.

\textit{MitoEM-R Cortex Data}\cite{wei2020mitoem} were scanned from Layer II/III in the primary visual cortex of an adult rat at a resolution of 8$\times$8$\times$30 $nm^3$. The publicly available MitoEM-R Data contain a training subset of 400$\times$4096$\times$4096 and a testing subset of size 100$\times$4096$\times$4096. For model training, we only use randomly selected 40 out of the 400  training images.

We evaluate our method under three  scenarios. First, we adapt from the small Drosophila  dataset to the medium-sized EPFL data. Second, we adapt from EPFL data to Kasthuri++ data, which contains a large proportion of background and a larger number of mitochondria. Third, we conduct the adaptation from EPFL dataset to the large MitoEM-R dataset.

\begin{figure}[t]
\centering
\includegraphics[width=0.48\textwidth]{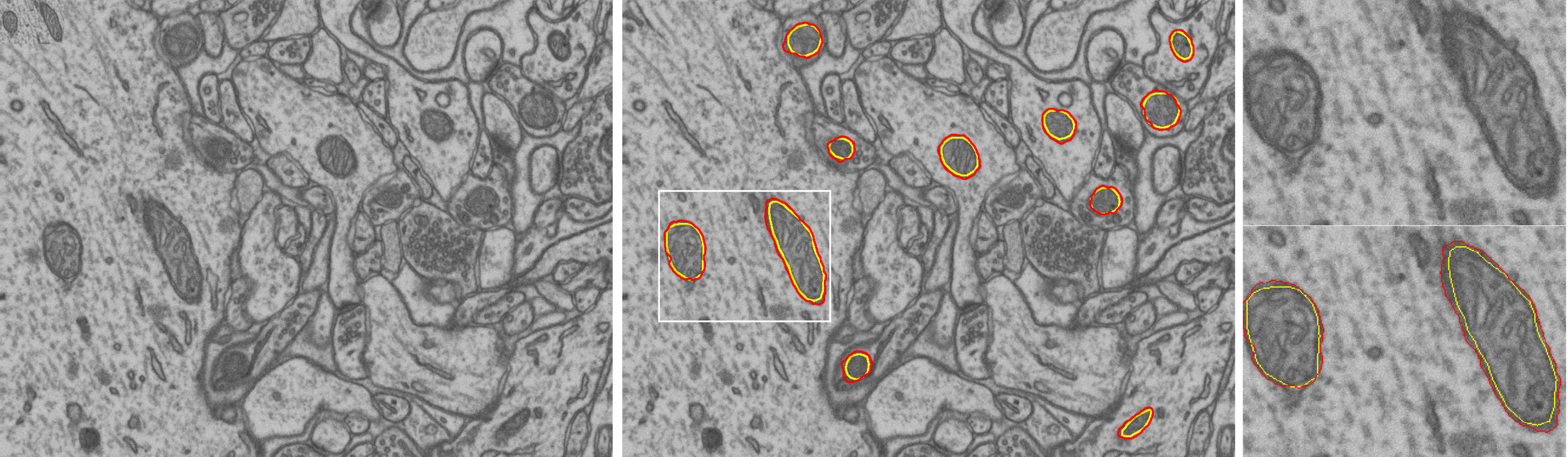}
\caption{Source annotation refinement (SAR) by GAC. Yellow: original annotation; Red: refined annotation by GAC. Note that the refinement is only used for source model training but not used for target data.   } \label{fig:5}
\end{figure}

\textbf{Source annotation refinement (SAR)}. While Mitochondria have a clearly visible double membrane as shown in Fig. \ref{fig:5},  the membrane was not included as the foreground in the annotations in EPFL data. In contrast, the membrane was annotated as part of the mitochondria in many other datasets, which introduces segmentation bias. Moreover, the EPFL labels contain more inconsistency on the boundaries. Thus, we introduce a source annotation refinement, which employs the Geodesic Active Contour (GAC)  \cite{Geodesic} to automatically refine the source labels while keeping target ground truth labels untouched. The GAC model can make the source annotations more consistent with image boundaries. An example is shown in Fig.\ref{fig:5}. The influence of the SAR will be evaluated in the following sections.

\textbf{Evaluation metrics.} Following previous studies\cite{peng2021csnet,yuan2021hive}, we  measure the segmentation performance with  Dice  coefficient (Dice), which is a class-level measure,  and  Aggregated Jaccard-Index (AJI) \cite{kumar2017dataset} and Panoptic Quality (PQ) \cite{kirillov2019panoptic}, which are  instance-level measures.

Dice is one of the most widely-used  criteria for medical image segmentation. Let $S$ and $G$ be the predicted binary segmentation  and ground truth annotation, respectively, Dice is defined as,
\begin{equation}
{\rm Dice} = \frac{2 |S\cap G| }{ |S| +|G|}
\end{equation}

Let  $G^j$ be the $j \rm th$ instance (i.e., mitochondrion) in $G$ with a total of
$N$ instances. Similarly, $S^j$ denotes as the $j \rm th$ instance in $S$. the AJI is defined as,
\begin{equation}
{\rm AJI} = \frac{\sum_{j=1}^{N} |G^j\cap S^{j^*}| }{\sum_{j=1}^{N} |G^j\cup S^{j^*}| +\sum_{i\in {\rm FP}} |S^{i}|},
\end{equation}
where $j^*$ is the index of the matched instance  in  $S$ with the largest overlapping with $G^j$; FP denotes false positive instances in $S$  without the corresponding ground truth mitochondria in $G$.

 The instance-level measure PQ is a hybrid measure of segmentation quality  in true positives (TP)  and detection quality, and is defined as,
\begin{equation}
\begin{matrix}
~~{\rm PQ} = \underbrace{\frac{\sum_{j\in {\rm TP}} {\rm JAC}(G^j, S^{j^*}) }{|{\rm TP}|}}&\times&\underbrace{\frac{{\rm |TP|}}{ {\rm |TP|}+\frac{1}{2}|{\rm FP}|+\frac{1}{2}|{\rm FN}|}}\\ \scriptsize \text{~~~~~~~~~~~~~~~~~~~Segmentation Quality}&&\scriptsize \text{Detection Quality}
\end{matrix}
\end{equation}
where FN denotes the set of false negatives.

\textbf{Network settings.} We use a lightweight variant of U-Net  as the backbone network of the  $\mathcal{G}_1$ network.
Instead of using standard 2D convolutions, the backbone network uses the HDD unit  in \cite{peng2021csnet} as the basic building blocks.
While one HDD layer is used in the segmentation head, the detection head contains two HDD layers. Given the heatmap predicted by the detection network, we use a small network comprised of one HDD  layer  and an integration layer to obtain the instance count $\hat{T}$.  For the counting network $\mathcal{G}_2$, we use the same backbone as the main network $\mathcal{G}_1$ and use an integration layer for counting prediction\cite{zhang2018crowd}. Followed \cite{peng2020unsupervised}, we set the discriminator network  as   a 5-layer fully-convolutional network  and set the channel
numbers of the 5 layers as \{64, 128, 256, 512, 1\}.

\textbf{Experimental  settings.} For model implementation, we set the parameters as $\lambda_a$=$10^{-3}$, $\lambda_d$=$10^{-1}$, and $\lambda$=$3$.  The proposed model is implemented with Pytorch and  trained on one GTX 1080Ti GPU with 11 GB memory. The main network $\mathcal{G}_1$ is optimized with stochastic gradient descent (SGD) using an  initial learning rate of $5\times 10^{-5}$ and batch size 2. Polynomial decay of power 0.9
 is used to control the learning rate decay. The maximum iteration number is set as 20k and the $z_{max}$ in consistency loss is set as 10k. For model training, the input images are randomly  cropped into smaller patches of size 512$\times$512. To improve model generalizability, we artificially increase data amount and data diversity with various  operations, such as blur, flipping, rotations,  and color jitters. Besides these general data augmentation operations,   the proposed CP-Aug as a specialized method for partial points is also used. Following \cite{Outputspace}, we train the discriminator with  Adam optimizer.
The counting network $\mathcal{G}_2$  initialized with parameters of the source detection network is optimized with mean squared loss and  Adam optimizer. To improve the cross-domain generalizability of the counting network, we use multi-scale inputs of size 512 $\times$ 512, 768 $\times$ 768, and 1024 $\times$ 1024 obtained through cropping and resampling as model inputs. Moreover, we use various operations including  flipping, rotations, blur, and color jitters as data augmentation to train the network $\mathcal{G}_2$.

\begin{table*}[t]
\caption{Ablation study of the proposed WDA-Net method on Drosophila III VNC $\rightarrow$EPFL Mouse Hippocampus and EPFL Mouse Hippocampus$\rightarrow$Kasthuri++.}
\centering
\label{tab:1}
\setlength{\tabcolsep}{3.2mm}
\begin{tabular}{lccccccccccc}\\
\hline
&&&&&&&\multicolumn{2}{c}{Drosophila $\rightarrow$EPFL}& ~&\multicolumn{2}{c}{EPFL $\rightarrow$Kasthuri++}\\
 \cline{8-9}\cline{11-12}
Model     &Detection & Count& Pseudo-label & CP-Aug& SAR&Filter & Dice(\%)  & PQ(\%)& ~  & Dice(\%)  & PQ(\%) \\\hline
I            & ~        & ~    & ~          & ~ & &       & 71.0& 45.6& ~ & 76.0& 49.6   \\
II       & \checkmark        & ~    & ~& ~   & ~  &        & 84.6& 60.8 & ~&86.3 &57.1    \\
III   & \checkmark        & \checkmark    & ~  & ~& ~   & & 85.9& 65.0 & ~ &88.7 & 68.3  \\
IV         & ~        & ~    & \checkmark & ~  & ~  &     &77.5  & 48.4& ~ &81.9 &55.2    \\
V      & \checkmark        & ~    & \checkmark  & ~& ~  & & 87.6 & 64.7  & ~& 89.0 &67.1  \\
VI & \checkmark& \checkmark    & \checkmark  & ~ & ~  &   & 89.3 & 70.4 & ~ & 89.8& 70.1  \\
VII  & \checkmark   & \checkmark    & \checkmark  & \checkmark && & 90.5 & 73.3 & ~&91.0 &  73.7 \\
VIII  & \checkmark  & \checkmark & \checkmark& \checkmark  &\checkmark& & 90.7 & 73.2 & ~ &93.2&   78.5 \\
WDA-Net  & \checkmark        & \checkmark    & \checkmark  & \checkmark & \checkmark  &\checkmark& \textbf{90.7}         & \textbf{76.5} & ~  &\textbf{93.2}& \textbf{80.2}  \\
\hline
\end{tabular}
\end{table*}

\subsection{Model Analysis}
\textbf{Ablation study}. As shown in Table \ref{tab:1},  we investigate   individual  contributions of the key components in our WDA-Net, a) Detection: using the detection task; b) Count: the proposed counting prior; c) Pseudo-label: the pseudo-label learning; d) CP-Aug:  the cross-position cut-and-paste label augmentation; e) SAR: the source annotation refinement with GAC; and f) Filter: refining  the segmentation from the detection head with connected component analysis  and also removing noise blobs with morphological operations. The ablations are conducted using WDA-Net (15\%), which  uses 15\% point annotation,  under two adaptation settings, i.e., Drosophila $\rightarrow$FPFL and EPFL $\rightarrow$Kasthuri++.

The Model I denotes the baseline model which is an UDA model with adversarial learning \cite{Outputspace}. As shown in Table \ref{tab:1}, sequentially adding the key components on top of the  Model I leads to  gradually improved performance. By integrating the center detection to the Model I, we obtain a large performance gain of 13.6\% in Dice, 15.2\% in PQ  for Drosophila $\rightarrow$FPFL and also a  substantial improvement of 10.3\% in Dice, 7.5\% in PQ for EPFL $\rightarrow$Kasthuri++. By integrating  the counting consistency  constraint, we get further performance gain, especially a large  performance gain  in detection, i.e., 4.2\% in PQ for Drosophila $\rightarrow$FPFL  and 11.2\% in PQ for EPFL $\rightarrow$Kasthuri++, which show the benefit of the global counting prior. By comparing the Model V,  IV, and  I, it can be seen that utilizing the Detection task and  the Pseudo-label learning can substantially improve both the segmentation  and  detection performance, while the performance gain in PQ by only using  Pseudo-label  is limited.  By comparing Model VII and  VI, it can be seen that the CP-Aug can  result in a performance gain of 2.9\% and 3.6\% in PQ for the two adaptation tasks, respectively. By comparing Model VIII and  VII, we observe that  CP-Aug is able to obviously improve the model performance for EPFL $\rightarrow$Kasthuri++ while having almost no influence  for Drosophila $\rightarrow$FPFL, because the EPFL data have obvious inconsistency in boundaries.
Finally, the detection outputs is able to help significantly reduce false positives,  and the full WDA-Net  outperforms Model VIII by 3.3\% and 1.7\% in PQ for the two adaptation tasks, respectively. In a nutshell,  all the proposed components can boost the model performance on different tasks, while the Detection, Count, CP-Aug, and Filter show relatively stronger ability than other key components to improve the detection performance.

\begin{figure}[t]
\centering
\includegraphics[width=0.33\textwidth]{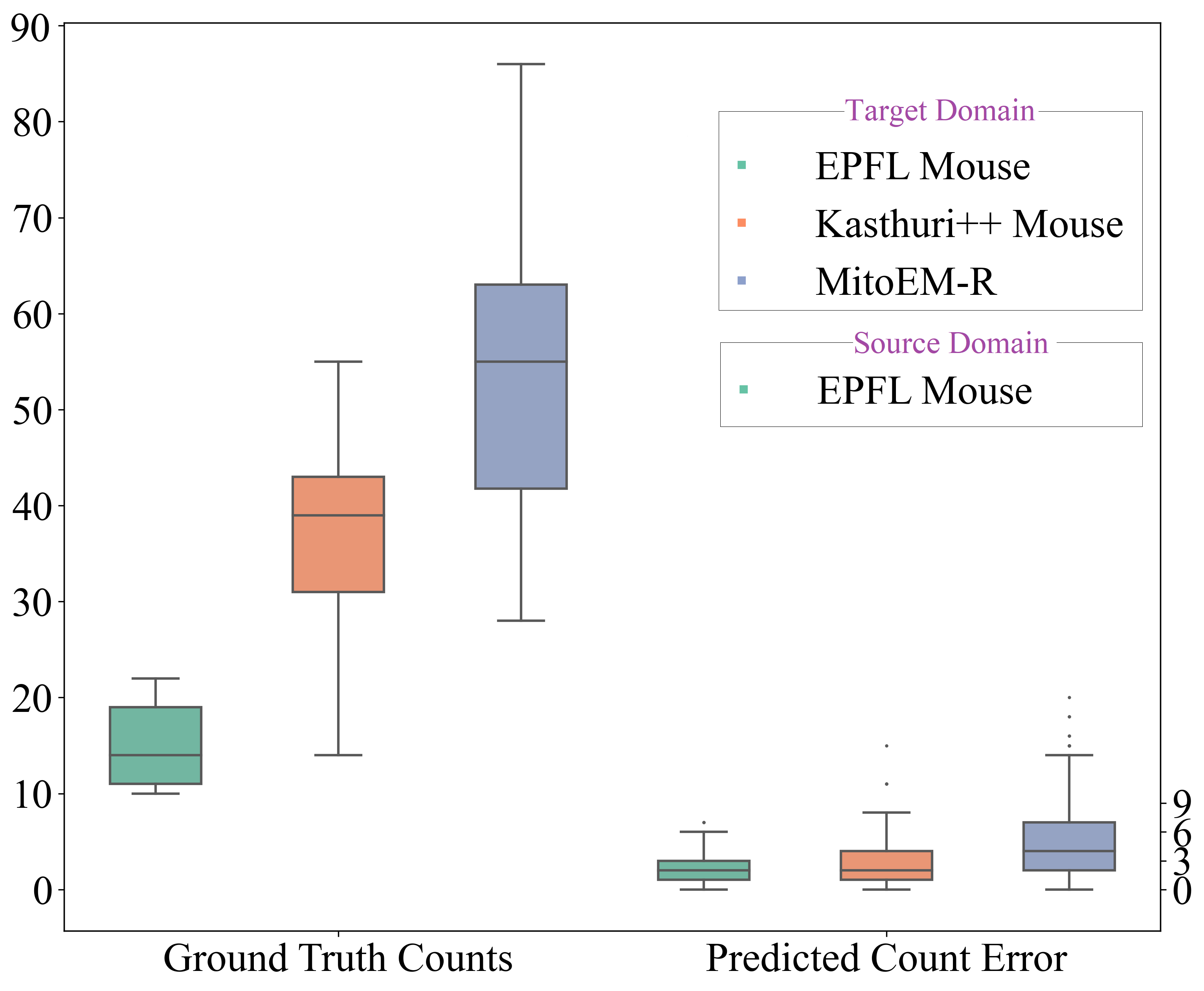}
\caption{Effectiveness of our counting network $\mathcal{G}_2$. The counting model is trained on EPFL training set and tested on EPFL testing set (images of size 1,024$\times$1,024), Kasthuri++ training set (images of size 1,334$\times$1,553), and MitoEM-R training set (cropped into images of size 1536$\times$1536). } \label{fig:6}
\end{figure}

\textbf{Effectiveness of the counting network.}
One critical challenge for the cross-domain counting task is the varied densities of target instances, which is alleviated with designed augmentations for better generalization  ability.  To validate the effectiveness of our $\mathcal{G}_2$  model as the counting prior, we test the prediction error of a counting model on images of similar sizes from different datasets. In the experiments of Fig. \ref{fig:6}, the counting model was trained on the EPFL training set and tested on the EPFL testing set, Kasthuri++ training set, and MitoEM-R training set. While the images from the EPFL testing set and Kasthuri++ training set are kept at their original sizes,  1,024$\times$1,024 and 1,334$\times$1,553 for the two datasets, respectively, the MitoEM-R images are cropped into smaller images of size 1,536$\times$1,536. As shown in Fig. \ref{fig:6}, despite  the varying number and appearance of the mitochondria across different domains, our counting model $\mathcal{G}_2$  demonstrates  constantly low counting error.
\begin{figure}[t]
\centering
\includegraphics[width=0.38\textwidth]{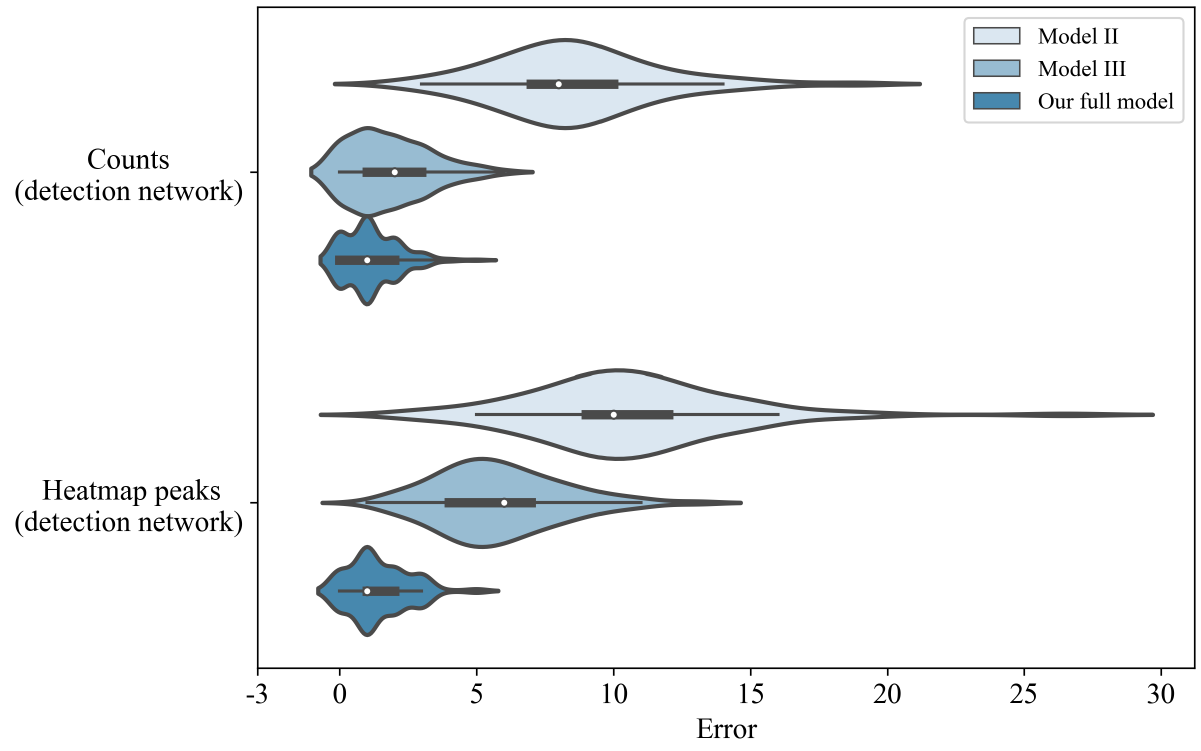}
\caption{Impact of the proposed counting-consistency constraint on counting  accuracy for Drosophila $\rightarrow$EPFL. The first group of violin plots  compares  the error distributions of the predicted counts $\hat{T}$, and the  second group compares the error distributions by counting local peaks (the top 80\%) in the predicted head maps $\hat{h}^t$.  } \label{fig:7}
\end{figure}

\begin{table}[t]
\caption{Robustness to different samples of sparse annotations (15\%). }
\centering
\setlength{\tabcolsep}{1mm}
\label{tab:2}
\begin{tabular}{lcccccc}
\hline
\multicolumn{7}{c}{Drosophila data $\rightarrow$EPFL data}                 \\
\hline
& Sample 1  &Sample 2 & Sample 3 &Sample 4&Sample 5 &Mean$\pm$Std   \\
\hline
Dice (\%)    & 90.7& 90.9 &90.4& 91.0& 91.1 &90.8$\pm$0.3    \\
PQ (\%)    & 76.5& 76.0 &75.7& 76.2 &78.9 &76.7$\pm$0.3\\
\hline
\end{tabular}
\end{table}

\textbf{Impact of the counting-consistency constraint.}
Figure \ref{fig:7} compares the distributions of the counting errors by the Model II, Model III, and our full WDA-Net.  Both the Model II and Model III use  adversarial learning on the segmentation head for model adaptation. While the detection task of Model II is only guided by the given sparse point annotations, the detection task of the Model III is also guided by the  proposed  counting consistency constraint. Note that  while enforcing supervision from the  sparse points to the detection head will partially guide the model training, the incomplete supervision also tends to  introduce false positive detection.  For comparison, we evaluate both the integral-based counting output $\hat{T}$ and the counted peaks (top 80\%) in the  predicted head map $\hat{h}^t$.  As shown in Fig. \ref{fig:7}, with the counting prior, Model III shows significantly lower counting errors than Model II, which validates the benefit of the proposed counting consistency constraint.

\begin{table*}[t]
\caption{Comparison results on Drosophila III VNC$\rightarrow$EPFL Mouse Hippocampus,  EPFL Mouse Hippocampus$\rightarrow$Kasthuri++, and EPFL Mouse $\rightarrow$MitoEM-R. For EPFL Mouse $\rightarrow$MitoEM-R, we use only 1/10 of MitoEM-R training set and 15\% sparse points are essentially about 1.5\% of the all center points. DA-ISC \cite{huang2022domain} is a 2.5D method that takes multi-slice input and uses inter-section consistency.  The Network parameters (M) and FLOPs (G) (with input size of 1024$\times$1024)
are reported. }
\centering
 \setlength{\tabcolsep}{0.5mm}
 \label{tab:3}
\begin{tabular}{lllcccccccccccccc}
\hline
&&&\multicolumn{3}{c}{Drosophila $\rightarrow$EPFL}& ~&\multicolumn{3}{c}{EPFL $\rightarrow$Kasthuri++}& ~&\multicolumn{3}{c}{EPFL $\rightarrow$MitoEM-R}&& GFLOPs&Params\\
\cline{4-6}\cline{8-10}\cline{12-14}\cline{12-14}\cline{16-17}
Type& Methods                & Backbone & Dice (\%) & AJI (\%)  & PQ (\%)& ~ & Dice (\%) & AJI (\%)  & PQ (\%)& ~ & Dice (\%) & AJI (\%)  & PQ (\%) & &(Inference)&M\\
\hline
 &NoAdapt                 &  U-Net   & 57.3        & 39.6        & 26.0  & ~ &70.0 & 52.8 & 42.2 &~&67.0&46.0&30.5 &&1076 &34.6  \\
 \hline
 \multirow{8}{*}{UDA}&Y-Net \cite{Y-Net}  &  \multirow{3}{*}{U-Net}    & 68.2        & --       & --  & ~  & 73.7 & 56.8 & 44.3   &~&72.8&50.9&35.1 && \multirow{3}{*}{1076}&\multirow{3}{*}{34.6}\\
&AdaptSegNet \cite{Outputspace}             &    & 69.9        & --       & --      & ~ &77.4 & 61.5 & 52.9 &~&74.9&55.3&43.6 && &\\
&DAMT-Net \cite{peng2020unsupervised}                &     & 74.7        & --        & --   & ~  & 82.0 & 68.3  &56.5  &~&80.4&62.8&54.7 &&&\\
\cline{2-17}
&Y-Net \cite{Y-Net}                   & \multirow{3}{*}{U-Net (HDD)}    & 69.6        & 52.2     & 42.6 & ~  & 76.7  &60.0 & 48.9 &~&74.9&56.5&40.7 && \multirow{3}{*}{74}& \multirow{3}{*}{3.7}\\
&AdaptSegNet \cite{Outputspace}             &   & 71.2       & 54.9     & 47.3  & ~ & 78.3 & 62.0  &50.7 &~&77.1&59.2&50.1  &&&\\
&DAMT-Net \cite{peng2020unsupervised}               &     & 75.3         & 59.7     & 47.7  & ~ &83.7  &70.2  &57.5  &~&83.4&67.9& 60.2&&& \\
\cline{2-17}
 &SAC \cite{araslanov2021self}                 & ResNet-101 & 77.6    & 63.6        &47.7         & ~  & 83.6 & 69.8  &50.1 &~&84.4&70.0& 55.4 &&723&42.7\\
&DA-ISC \cite{huang2022domain} (2.5D)                  & Two-head U-Net & 81.3    & 68.2        &60.0         & ~  & 85.2 & 73.0  &63.2 &~&78.5&61.5& 51.8 &&1258&15.3\\
\hline
 \multirow{4}{*}{WDA}&Our model (5\%)      &  \multirow{4}{*}{U-Net (HDD)}
                            & 88.5& 79.3 &74.5    & ~&92.6 &84.7 &78.0 &~&88.8&75.7&71.7&&  \multirow{4}{*}{86}& \multirow{4}{*}{3.8}\\
&Our model (15\%)      &    & 90.7& 82.1& 76.5  & ~ & 93.2& 86.2 &80.2     &~&90.1&77.5& 73.1 &&&\\
&Our model (50\%)       &   & 91.0& 83.4 &77.8  & ~ &93.5& 86.4 &80.5 &~&91.4&79.1&73.9 &&& \\
&Our model (100\%)      &  & 91.6& 83.9 &78.3 & ~  & 94.3 &87.8 &82.1   &~&92.0&80.8&75.6  &&&\\
\hline
Oracle &       Supervised model     &  U-Net (HDD)  & 93.6       & 87.9     & 80.2 & ~ & 94.6 &88.3 &82.2 &~&93.6&83.8&77.1 &&74&3.7 \\
\hline
\end{tabular}
\end{table*}

\textbf{Robustness to different annotations.} Robustness to annotation selections is important for the practicality of weakly-supervised models. To this end, this study also investigates the  performance of our model with different random selections of 15\% sparse annotations on the adaptation from Drosophila data to EPFL data. As demonstrated in Table \ref{tab:2}, the proposed WDA-Net shows robust
performance with small performance variance, which is highly desirable in practical applications.

\begin{figure*}[t]
\centering
\includegraphics[width=0.8\textwidth]{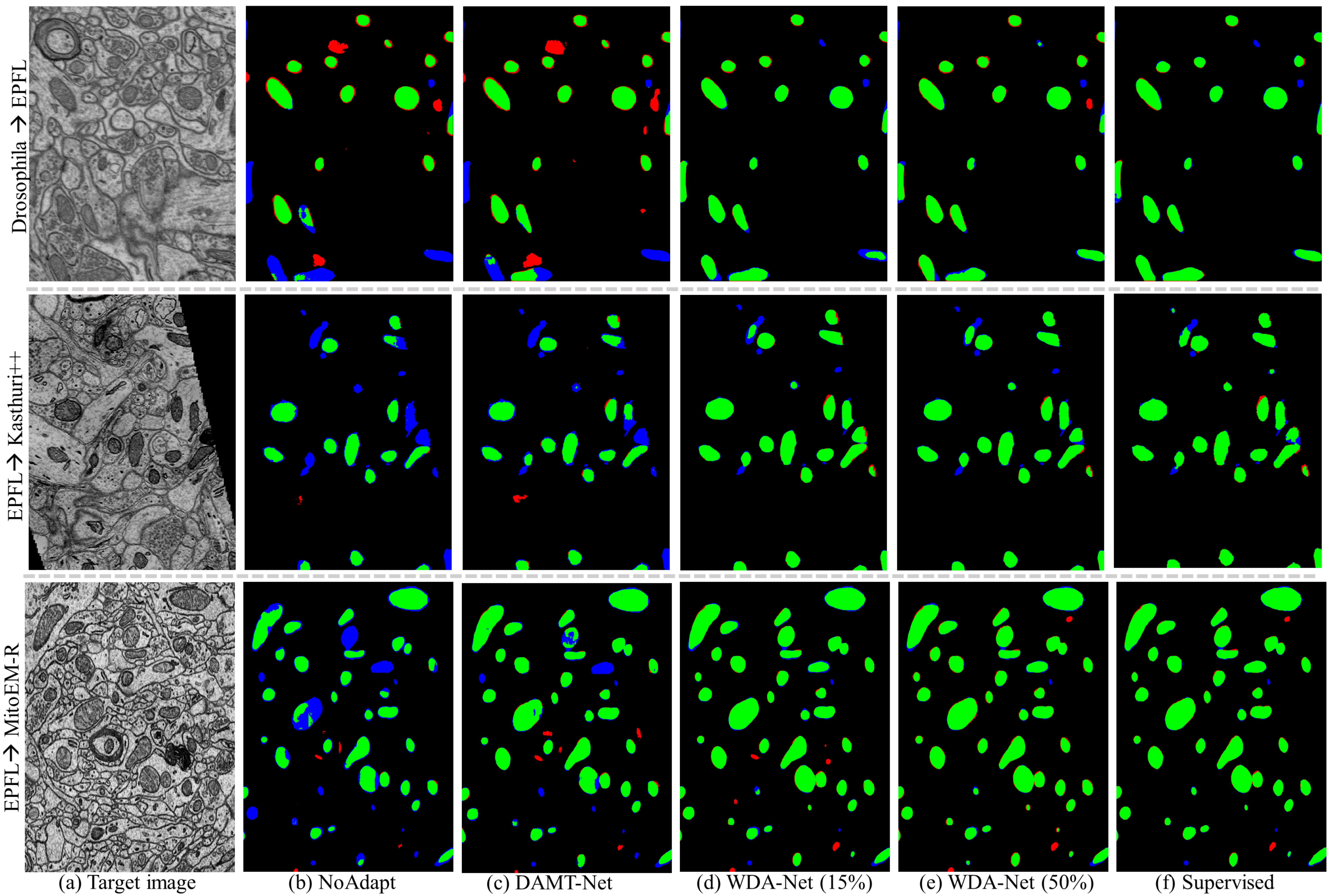}
\caption{Segmentation results of different methods. Green:  true positives; Red:  false positives;  Blue: false negatives.  } \label{fig:8}
\end{figure*}

\subsection{Comparative Experiments}
In Table \ref{tab:3}, we  quantitatively compare our WDA-Net with SOTA UDA models, i.e., Y-Net \cite{Y-Net}, AdaptSegNet \cite{Outputspace},  DAMT-Net \cite{peng2020unsupervised}, SAC \cite{araslanov2021self}, DA-ISC \cite{huang2022domain},  the upper bound, i.e., the supervised trained model on the target domain, as well as the lower bound, i.e., the source models without adaptation. Both the  U-Net \cite{ronneberger2015u} and its lightweight variant U-Net (HDD), which uses the HDD \cite{peng2021csnet} module as the basic building blocks, are investigated as the backbone network. In SAC \cite{araslanov2021self}, the DeepLab V2 \cite{chen2017deeplab} with pretrained ResNet-101 \cite{he2016deep} as its backbone was used for segmentation. Note that DeepLabV2 (ResNet101) is a large model with 42.7M parameters, while the U-Net (HDD) has only 3.7M parameters. DA-ISC \cite{huang2022domain} is a 2.5D model that uses multi-slices as input and takes advantage of inter-section consistency. Thus, DA-ISC uses a variant of U-Net  with
three decoders. To identify the influence of annotation amount, we also compare  the performance of our proposed WDA-Net  when using various ratios of point annotations in Table \ref{tab:3}.

Firstly, it can be seen that models using the standard U-Net (34.6M) obtain lower performance than that our proposed U-Net (HDD)  with  only about 3.7M learnable parameters. Therefore,  U-Net (HDD) is chosen  as the default backbone network in all experiments. Moreover, as shown in the last two columns of Table \ref{tab:3}, our method with  U-Net(HDD)  takes significantly lower computation costs and lower GPU memory usage for parameter storing during model inference than other methods such as SAC and DA-ISC. For example, while the FLOPs (computed with an input image size of 1024$\times$1024 ) and Params of the SAC method are 723G and 42.7M, respectively, the FLOPs and Params of our WDA-Net are only 86G and 3.8M respectively, which indicates the efficiency of our method.

Secondly, for both adaptation tasks, we have observed significant degradation in segmentation performance when directly applying the source models  to the target domains. These results also indicate the existence of the severe domain shift and the sensitivity of deep convolution networks to shifted data distributions. By leveraging the unlabeled target data, the UDA methods demonstrate substantially improved performance over the NoAdapt. However, the performance of the UDA methods is still significantly lower than the supervised model and far from practical usage.

Thirdly, with minimal annotation effort, our  WDA-Net  significantly outperforms SOTA  UDA methods including the 2.5D model DA-ISC  on all three adaptation tasks in terms of both class-level and instance-level metrics, and constantly achieves comparable  performance as the supervised model. By comparing the performance of our model with different annotation ratios, it can be seen that sparsely annotating  15\% of instances in target training images is able to train a model having strong testing performance. For the Drosophila III VNC$\rightarrow$FPFL Mouse Hippocampus, the proposed WDA-Net with only 15\% sparse points is able to achieve substantial improvements over  the DAMT-Net, SAC, and  DA-ISC,  which are top-performing UDA methods. Compared to the supervised model that is trained with all the labels of the training images, the proposed method  requires significantly reduced annotation cost and expert knowledge (see Table \ref{tab:4}), which is quite favorable for practical usage. Our model also shows greatly improved results in the other two more challenging settings, especially in terms of instance-level measures. For  EPFL$\rightarrow$Kasthuri++, our WDA-Net (15\%) achieves a performance of 86.2\% in AJI, 13.2\% higher than the 2.5D model DA-ISC and only 2.1\% lower than the upbound, and 93.2\% in DSC, 8.0\% higher than DA-ISC and only 1.4\% lower than the upbound. For EPFL$\rightarrow$ MitoEM-R, our model uses only 1/10 of the MitoEM-R training data for training, thus the WDA-Net (15\%) essentially uses only about 1.5\% of full points for supervision. However, our WDA-Net (15\%) achieves a performance of 90.1\% in Dice, 11.6\% higher than DA-ISC, and 3.5\% lower than the upbound, which further confirms the benefit of our WDA-Net.

Figure \ref{fig:8} provides a visual comparison of the  WDA-Net with other methods under different adaptation settings. Without model adaptation, the NoAdapt results in a lot of false positives and false negatives on both segmentation tasks. Using unsupervised adaptation, the DAMT-Net in Fig. \ref{fig:8} (c) shows improved segmentation, but the segmentation still has many false positives and false negatives. In contrast, with minimal label cost, the proposed WDA-Net shows greatly reduced false positives and false negatives and shows comparable results as the fully-supervised counterpart in Fig. \ref{fig:8} (f). Compared to the setting with 50\% center points, the WDA-Net using 15\% center points on the training data can already dramatically outperform state-of-the-art UDA  methods.

\begin{table*}[t]
\caption{Annotation time and accuracy of four non-expert annotators. The EPFL testing data containing 165 slices and over 2600 2D instances are used for annotation. The Round II of full point annotation is designed for annotation revision.}
\centering
 \setlength{\tabcolsep}{2mm}
 \label{tab:4}
\begin{tabular}{cccccccccccc}
\hline
EPFL Testing set &\multicolumn{2}{c}{Sparse point (15\%)}& ~&\multicolumn{2}{c}{Full point (Round I)}& ~&\multicolumn{2}{c}{Full point (Round I+II)}& ~&\multicolumn{2}{c}{Full pixel-wise label}\\
\cline{2-3}\cline{5-6}\cline{8-9}\cline{11-12}
(165 slices) & Time (min) & Recall (\%) & ~ &Time (min) & Recall (\%) & ~ & Time (min) & Recall (\%) & ~ &Time (min) & DSC(\%)  \\
\hline
Annotator 1& 	16.5& 	100& 	& 75.5& 	81.1& 	& 90.0& 	85.6& 	& 330.0& 	84.5\\
Annotator 2& 	12.0& 	100& & 	58.3& 	85.2& 	& 78.3	& 89.2& 	& 363.0	& 88.5\\
Annotator 3	& 11.5	& 100& 	& 60.0& 85.4&  & 79.5 &90.9& 	& 291.5	& 88.0\\
Annotator 4& 	18.5& 	100	& & 73.5	& 88.6& 	& 110.5& 	92.8& 	& 306.9& 	90.0\\\hline
Mean& 	14.6& 	100	& & 66.8& 	85.1& & 	89.5& 	89.6& & 	322.9& 	87.8\\
 \hline
\end{tabular}
\end{table*}

\subsection{Visualization of Aligned Feature }
\begin{figure}[t]
\centering
\includegraphics[width=0.45\textwidth]{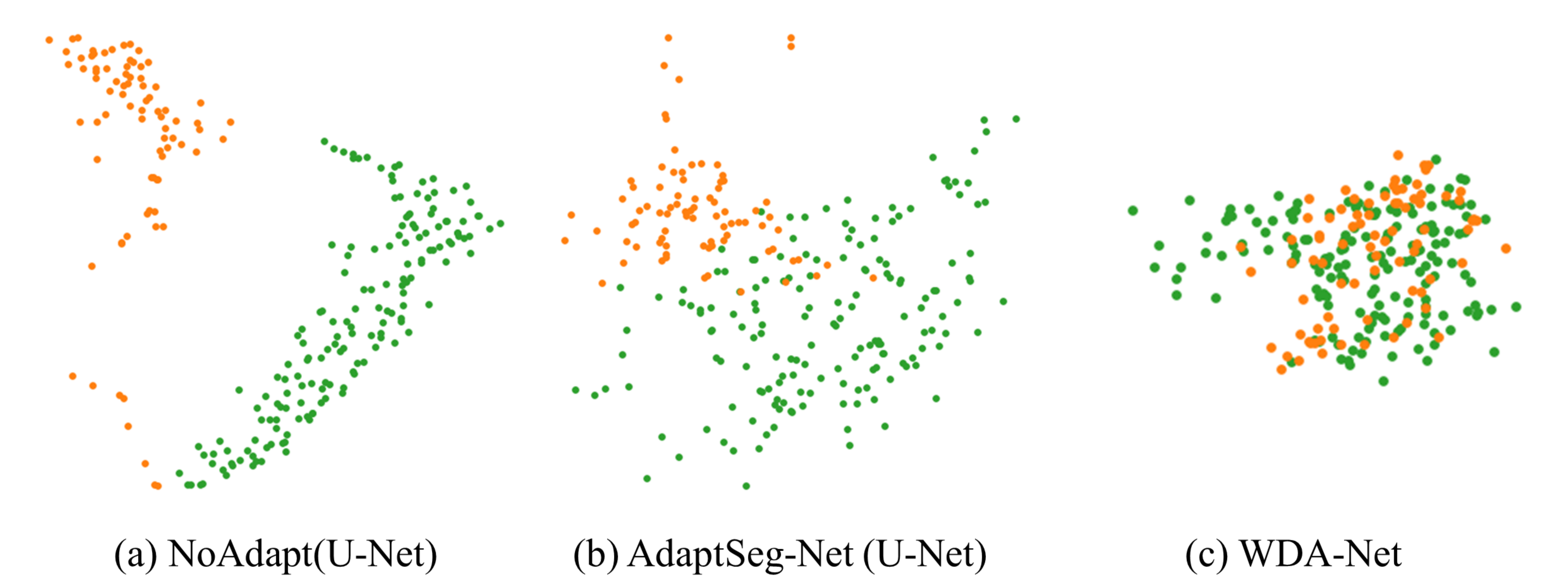}
\caption{Visualization of features  with t-SNE before and after domain
adaptation from EPFL data to Kasthuri++ data. Different domains are denoted with different colors. } \label{fig:9}
\end{figure}

Figure \ref{fig:9} visualizes the  features of the penultimate layer before and after domain adaptation with two methods, the AdaptSegNet (U-Net) and our WDA-Net (15\%). After the adaptation from EPFL data to
 Kasthuri++ data, our model outputs more aligned  features, which demonstrates the effectiveness of our domain adaptation method.

\subsection{Annotation Efficiency}
Compared to full center-point annotation and full pixel-wise annotation illustrated in Fig. \ref{fig:2}, our sparse center-point annotation not only takes a greatly reduced annotation workload but also requires much less expert knowledge, since the annotator just needs to annotate several most easily identifiable instances in each slice/block. Thus, non-expert annotators just having some knowledge about EM images can be employed for data annotation, which is very valuable for practical usage.

 To quantitatively demonstrate the annotation efforts of different annotation types, e.g., sparse point, full center-point, and full pixel-wise label, we employ four non-expert annotators for manual annotation. Annotators 1, 2, and 3 have little knowledge about the EM images and also reviewed the ground truth segmentation  3-5 times before the annotation. Annotator 4 is more familiar with EM images and working on EM image analysis. The EPFL testing data, which contains 165 2D slices and over 2600 2D instances, is used for comparison.

For the sparse point annotation, the annotators are required to annotate 2-4 instances in each slice. For full point annotation, the annotators are required to annotate for two rounds, and the second round is for revision after another review of the ground truth segmentation. For full pixel-wise annotation, to accelerate the annotation, the annotators are required to draw the boundaries of the object instances and then the hole-filling is used to automatically transform the boundary annotation into pixel-wise labels. The label transformation time is not computed as the annotation time. No second-round revision is conducted for pixel-wise annotation.

  As shown in Table \ref{tab:4}, it is challenging for non-expert annotators to correctly identify all target instances. For the full center-point annotation, the four annotators show a mean Recall (TP/(TP+FN)) of 85.1\% for the Round I annotation, and the final Recall after revision is 89.6\%. The annotators take about 67 minutes to complete the Round I annotation and about 90 minutes to complete the two rounds of annotation. In contrast, the annotators take only about 15 minutes to complete the sparse-point annotation with a Recall of 100\%. For the full pixel-wise annotation, the mean annotation accuracy in terms of Dice is only 87.8\%, lower than 90\%, and the mean annotation time is 322.9 minutes, which is 21.5 times of the proposed sparse-point annotation.

\section{Conclusions and Discussions }\label{sec:conclusion}
This study devises  a task-pyramid learning framework to tackle the performance limitations of domain-adaptive segmentation of EM images under extremely weak supervision. To conduct  segmentation at minimal annotation cost and minimal expert knowledge,  we introduce  sparse point labels on partial mitochondria instances in the training EM images. Given the incomplete supervision, we conduct multi-level domain alignment by taking advantage of  the joint learning of  segmentation, center detection, and counting, which are correlated tasks with different levels of domain invariance. More specifically, counting is introduced as a soft global prior for  center detection, which is modeled as a location regression  under partial annotations. We also introduce a novel cross-position  cut-and-paste strategy to further alleviate  the label sparsity. Extensive validations and ablation studies on multiple challenging benchmarks have demonstrated the effectiveness and robustness  of our WDA-Net, which can obtain  performance close to the supervised model with only 15\% sparse point supervision.

Despite the promising performance, this study has two main limitations.
First, the same as  most domain adaptation methods,  it assumes the availability of  both the  source model and the well-labeled source  data, which  may not always be  feasible in clinical applications due to privacy concerns. Thus, as a future study, we will investigate the setting without access to source data and conduct model adaptation with only the source model and  target data.
Second, the proposed model utilizes counting/detection as the auxiliary tasks, which is especially  suitable for segmentation objects with many instances, such as cellular segmentation. However, the proposed sparse point annotation and the WDA-Net are not  feasible for standard   semantic segmentation tasks, such as organ segmentation from abdominal medical images.   While we focus on EM image segmentation in this study,  we will validate our method on more cellular segmentation tasks as future work.


\vfill

\end{document}